\documentclass{article}
\usepackage{graphicx} 
\usepackage{authblk}
\usepackage{amssymb}
\usepackage{graphicx}
\usepackage{xcolor}
\usepackage{caption}
\usepackage{multirow}
\usepackage{array}
\usepackage{subcaption}
\usepackage{natbib}
\usepackage{hyperref}

\title {YOLO-ELA: Efficient Local Attention Modeling for High-Performance Real-Time Insulator Defect Detection}
\author{Olalekan Akindele$^{1,2*}$ and Joshua Atolagbe$^{2*}$}
\affil{$^1$DAIM, University of Hull, UK; $^{2}$ OOD Technologies, UK.
$^{*}$Equally contributing first authors}
\date{}
\begin{document}

\maketitle
\begin{abstract}
\noindent Existing detection methods for insulator defect identification from unmanned aerial vehicles (UAV) struggle with complex background scenes and small objects, leading to suboptimal accuracy and a high number of false positives detection. Using the concept of local attention modeling, this paper proposes a new attention-based foundation architecture, YOLO-ELA, to address this issue. The Efficient Local Attention (ELA) blocks were added into the neck part of the one-stage YOLOv8 architecture to shift the model's attention from background features towards features of insulators with defects. The SCYLLA Intersection-Over-Union (SIoU) criterion function was used to reduce detection loss, accelerate model convergence, and increase the model's sensitivity towards small insulator defects, yielding higher true positive outcomes. Due to a limited dataset, data augmentation techniques were utilized to increase the diversity of the dataset. In addition, we leveraged the transfer learning strategy to improve the model's performance. Experimental results on high-resolution UAV images show that our method achieved a state-of-the-art performance of 96.9\% mAP$_{0.5}$ and a real-time detection speed of 74.63 frames per second, outperforming the baseline model. This further demonstrates the effectiveness of attention-based convolutional neural networks (CNN) in object detection tasks.
\end{abstract}

\section{Introduction}
\noindent The reliable operation of transmission power line infrastructure is vital to ensuring a stable electricity supply, meeting the energy needs of both individuals and businesses. As such, the inspection and maintenance of transmission tower components, like insulators, for defects is critical for ensuring the safe functioning of power grid systems. Insulators, which provide insulation for conductors and support cables, are susceptible to damage from harsh weather conditions or electromechanical stress \citep{sanyal2020failure}. This can disrupt the smooth operation of transmission networks, making routine inspection and maintenance necessary to identify and replace damaged insulators. Manual detection methods usually involve visual inspection of power lines by tower personnel. However, this method is labour-intensive due to the vast number of transmission towers and the distances between them, and it poses safety risks as workers often have to climb tall towers \citep{wei2024insulator}. A semi-automated alternative involves analyzing images taken by UAVs or helicopters using traditional image processing algorithms (\citealp{li2010recognition}; \citealp{wu2012texture}; \citealp{zhai2017fault}). However, due to the large number of high-resolution images being processed, these methods take time and often lead to misinterpretation errors \citep{liu2023a}. Moreover, these methods are limited by their sensitivity to complex backgrounds and experience difficulty in identifying small insulator defects (\citealp{wen2021defect}; \citealp{cheng2022image}; \citealp{liu2023b}). This prompts the urgent need for a fully automated solution.  \\

\noindent In the past decades, computer vision and deep learning methodologies have been increasingly used to automate various object detection tasks. The widespread adoption of neural network architectures, particularly Deep Convolutional Neural Networks (DCNNs), has led to significant improvements in both accuracy and speed over traditional detection methods. These gains are attributed to their ability to extract and learn high- and low-level features from image data like insulator datasets. In addition, they benefit from transfer learning strategies, which enhance performance by leveraging pre-trained weights \citep{liu2021improved}. \\

\noindent Current deep learning-based research on insulator defect detection uses two categories of DCNN detectors. One includes popular two-stage detector algorithms like R-CNN \citep{girshick2014rich}, Fast-RCNN \citep{girshick2015fast}, and Faster-RCNN \citep{ren2016faster}, which operate on the principle of candidate region proposal, followed by refining and identification of defect regions. For instance, \cite{wen2021defect} proposed two Faster R-CNN-based methods, Exact R-CNN (exact region-based convolutional neural network) and CME-CNN (cascade the mask extraction and exact region-based convolutional neural network), which incorporate advanced techniques like FPN, Generalized IoU (GIoU), and mask extraction to improve insulator defect detection accuracy in complex backgrounds and small targets reaching up to 88.7\%. Similarly, \cite{tang2022insulator} implemented an improved Faster R-CNN model for insulator defect detection in UAV aerial images by replacing VGGNet16 with ResNet50, incorporating a feature pyramid network (FPN; \citealp{lin2017feature}) for feature fusion, and using RoIAlign network to minimise quantization effects resulting in 84.37\% detection accuracy. Although these methods deliver high accuracy in challenging scenes, their deep networks lead to slower processing speeds, often falling short of real-time detection requirements of 40 frames per second \citep{chen2023a}.\\

\noindent To address this limitation, one-stage detectors like SSD (Single Shot MultiBox Detector; \citealp{liu2016ssd}) and YOLO (You Only Look Once; \citealp{jocher2020yolov5}; \citealp{jocher2023ultralytics}; \citealp{wang2023yolov7}) series have emerged as popular alternatives. These models significantly improve detection speed while maintaining high accuracy, making them more suitable for real-time applications. For instance, \cite{adou2019insulator} used YOLOv3 to detect insulators and identify defective ones, achieving up to 45 frames per second (FPS), which meets real-time detection requirements. Similarly, \cite{li2022insulator} introduced a YOLOv5-based method for fast and accurate insulator and defect detection in transmission lines, achieving 97.82\% accuracy with real-time detection at 43.2 FPS. This further highlights the high precision and speed of YOLOv5 in detecting insulator damage in complex environments, while its lightweight architecture makes it ideal for UAV deployment, improving inspection efficiency. However, \cite{ding2022high} pointed out that baseline YOLOv5 can still be affected by background interference due to its anchor settings, resulting in false positive detection. Furthermore, the distance of UAV aerial photography often results in insulator defect targets being detected with minimal pixel information in the images \citep{hu2023multi}. This prompted the increased adoption of the anchorless YOLOv8, featuring enhanced architecture that includes attention modules to improve defect detection accuracy and speed (\citealp{chen2023b}; \citealp{li2024location}; \citealp{su2024insulator}; \citealp{zhang2024improved}). Integrating attention modules in the convolution-based YOLOv8 model aims to shift the model’s attention from learning general features to features specific to insulators with defects, resulting in higher true positive predictions.  \\

\noindent In this paper, we propose a novel YOLOv8 architecture based on the Efficient Local Attention (ELA; \citealp{xu2024ela}) module to enhance both the accuracy and speed of insulator defect detection in high-resolution UAV aerial images. In this paper:
\begin{enumerate}
    \item We integrated the base variant of ELA into the neck component of YOLOv8 architecture to localize features related to insulators with defects. 
    \item In addition to ELA, we tested other attention modules like Convolutional Block Attention Module (CBAM; \citealp{woo2018cbam}), Efficient Channel Attention (ECA; \citealp{wang2020a}, Coordinate Attention (CA; \citealp{hou2021coordinate}), and Mixed Local Channel Attention (MLCA; \citealp{wan2023mixed}). 
    \item We implemented the SIoU (\citealp{gevorgyan2022siou}) criterion function to reduce prediction loss and higher true positive detection in small-pixel information scenarios. 
    \item We conducted an ablation experiment to compare the performance of the baseline YOLOv8 with the improved architecture.  
\end{enumerate}

\section{YOLOv8 Baseline Architecture}
\noindent YOLOv8 (\citealp{jocher2023ultralytics}) is a state-of-the-art convolution-based vision foundation model used for various computer vision problems, including object detection. It has up to five variants with the same architecture but differentiated by their number of parameters, overall performance, and computational demand. The larger variants yield more performance at the expense of computational loads. In this research, we adopted the smaller variant YOLOv8s (small). The architecture of YOLOv8s like other YOLOv8 variants, is defined by backbone network, neck, and head components. The backbone network consists of convolutional modules and the C2f (Cross Stage Partial with two fusion; \citealp{wang2020b}) modules, which itself is based on the C3 module from YOLOv5 \citep{jocher2020yolov5} and the Extended ELAN (Efficient Layer Aggregation Network; \citealp{wang2022gradient}) from YOLOv7 \citep{wang2023yolov7}. The C2f consists of two convolutional modules with multiple Darknet bottlenecks. These function as feature extractors, with the C2f module reducing computational complexity by splitting and concatenating channel dimensions. The backbone is connected to the neck component via the Spatial Pyramid Pooling Fast (SPPF) layer. The neck, acting as a bridge between the backbone and the head, incorporates a hybrid of the PAN (Path Aggregation Network; \citealp{liu2018path}) and the FPN, allowing it to capture rich feature maps, which are then passed to the decoupled head module that contain the classification and detection branches for final bounding box prediction.
\section{Improved YOLOv8 Architecture}
Recent advancements in convolution-based foundation models have introduced the concept of attention mechanism, an idea originally developed in transformer-based models, to enhance performance and accuracy. This includes spatial attention, which is designed to learn pixel-wise spatial information, and channel attention, which focuses on channel-wise dependencies. Integrating these attention mechanisms within convolutional blocks can lead to more robust feature representations, disregarding non-salient information and ultimately improving detection accuracy. In this work, we aim to improve how the rich features related to insulators with defects are captured by introducing the ELA \citep{xu2024ela} module in the neck component of YOLOv8. Furthermore, we adopted the SIoU criterion loss function to improve the model’s convergence and detection accuracy on small insulators with defects.
\subsection{Efficient Local Attention (ELA)}
Existing attention modules like CBAM \citep{woo2018cbam} and CA \citep{hou2021coordinate} lack sufficient generalization ability from using batch normalization, fail to capture long-range dependencies and reduce the channel dimensions of feature maps. In contrast, the ELA block aims to leverage robust spatial information without reducing channel dimensions or increasing complexity, aiding DCNNs in accurately localizing objects of interest. This block uses strip pooling \citep{hou2020strip} instead of spatial global pooling, an idea from CA, in the spatial dimension to obtain rich feature vectors to capture long-range dependencies in both horizontal and vertical directions. This ensures only features corresponding to the target regions are retained, disregarding irrelevant region features. A 1D convolution is then used for faster and lightweight local processing of the feature vectors from each direction, with a kernel resizing option to control the scope of local interaction. Group Normalization (GN; \citealp{wu2018group}) and non-linear activation function then refine the resulting feature maps to produce the final positional attention prediction from both directions. This significantly improves the overall performance and generalization of CNN-based models with only a slight increase in parameters. \\

\begin{figure}[htp]
    \centering
    \includegraphics[width=\textwidth]{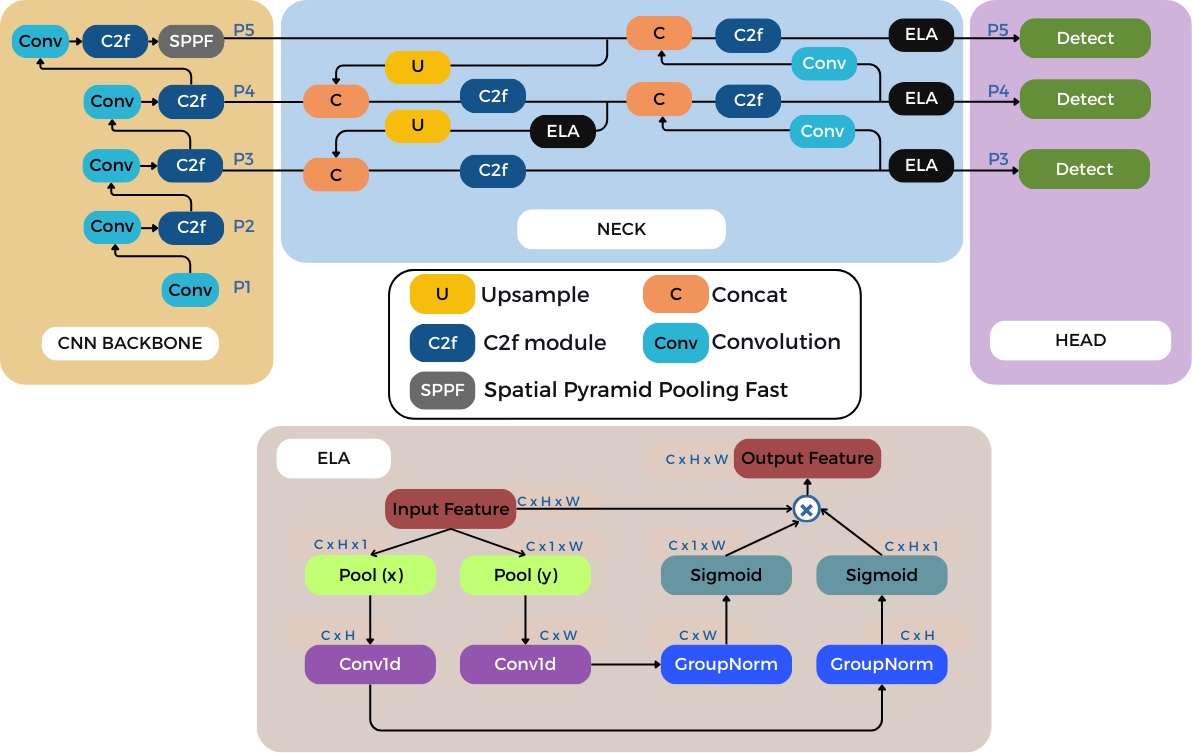}
    \caption{The Proposed YOLO-ELA Architecture}
    \label{fig:Figure 1}
\end{figure}

\noindent In this study, we modified the YOLOv8 architecture by incorporating the ELA blocks into the neck component after each of the four C2f modules (Cross Stage Partial bottlenecks with two convolutions) to enhance insulator defect detection in high-resolution UAV aerial images (See \hyperref[fig:Figure 1]{Figure \ref*{fig:Figure 1}}). Strip pooling is applied to the C2f output $x_c \in$ $\mathbb{R}^{H\times W\times C}$ (representing the height, width, and channel dimensions) across each channel along the horizontal $(H,1)$ and vertical $(1,W)$ directions, generating a representation for the \emph{c-th} channels at height $h$ and width $w$. \\
$$z_c^h(h) = \frac{1}{H} \sum_{0 \leq i < H} x_c(h, i)$$             
$$z_c^w(w) = \frac{1}{W} \sum_{0 \leq j < W} x_c(j, w)$$\\
Two feature maps are derived from these bidirectional channels and processed using two 1D convolutions $F_h$ and $F_w$ and Group Normalization $G_n$ (\emph{number of groups = 16}) to enhance and process spatial information to generate positional attention maps for both the horizontal $y^h$ and vertical $y^w$ directions. The sigmoid $\sigma$ activation function was used to apply a non-linear transformation to the maps with a convolutional kernel size of 7. 
$$y^h = \sigma(G_n(F_h(z_h)))$$
$$y^w = \sigma(G_n(F_w(z_w)))$$
The output $Y$ of the ELA block is a local attention map derived from multiplying the C2f features and attention maps of both directions, which captures the refined spatial information for accurate insulator defect detection. 
$$Y = x_c \times y^h \times y^w$$
\subsection{SIoU Loss Function}
The IoU loss metric is used in object detection tasks to measure the degree of overlap between the predicted and target bounding boxes. YOLOv8, by default, uses a fusion of the Distribute Focal Loss (DFLoss) and Complete Intersection over Union (CIoU) Loss in the regression branch to reduce detection loss during training. While CIoU considers factors like bounding box overlap, central point distance, and aspect ratio, it does not account for the trajectory of mismatch between regression boxes. This limitation may result in slower convergence and poor model performance. In this study, we replace CIoU with the SIoU (\citealp{gevorgyan2022siou}) loss to overcome this limitation. The SIoU criterion function enhances model convergence and performance by integrating four losses: angle cost, distance cost, shape cost, and IoU cost, providing a more robust evaluation of the bounding box mismatch and improving detection accuracy.\\
\begin{figure}[htp]
    \centering
    \includegraphics[width=8cm]{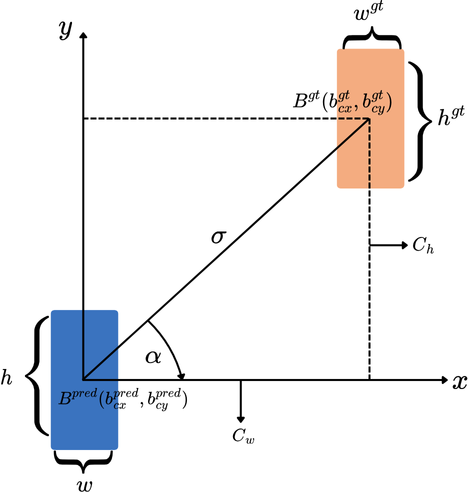}
    \caption{The Schematic Diagram of SIoU Loss}
    \label{fig:Figure 2}
\end{figure}

\noindent \hyperref[fig:Figure 2]{Figure \ref*{fig:Figure 2}} shows the schematic representation of the SIoU loss function. $B^{pred}$ and $B^{gt}$ represent the center point positions of the predicted and ground-truth bounding boxes, respectively. The coordinates of these center points are denoted as ($b^{gt}_{cx}$, $b^{gt}_{cy}$) for the ground-truth box and ($b^{pred}_{cx}$, $b^{pred}_{cy}$) for the predicted box. The angle and distance between the central points $B^{pred}$ and $B^{gt}$ are defined by $\alpha$ and $\sigma$, respectively. $C_w$ and $C_h$ represent the differences between $B^{pred}$ and $B^{gt}$ in the horizontal and vertical coordinates, respectively. At the same time, $(w, h)$ and $(w^{gt}, h^{gt})$ denote the width and height of the predicted and ground-truth boxes, respectively. The parameter $\theta$ adjusts the emphasis on the shape loss during training. We represent the loss functions in mathematical expressions below.\\

The angle loss $\Lambda$ is defined as:
$$\Lambda = 1 - 2 \sin^2 \left( \arcsin(x) - \frac{\pi}{4} \right)$$
$$x = \frac{c_{h}}{\sigma} = \frac{\max\left(b_{cy}^{gt}, b_{cy}^{pred}\right) - \min\left(b_{cy}^{gt}, b_{cy}^{pred}\right)}{\sqrt{\left(b_{cx}^{gt} - b_{cx}^{pred}\right)^2 + \left(b_{cy}^{gt} - b_{cy}^{pred}\right)^2}} = \sin(\alpha)$$

The distance loss $\Delta$ is defined based on the angle loss $\Lambda$ and represented as:
$$\Delta = \sum_{t=x,y} (1 - e^{-(2-\Lambda)\rho_t})$$
$$\rho_x = \left(\frac{b_{cx}^{gt} - b_{cx}^{pred}}{C_w}\right)^2, \rho_y = \left(\frac{b_{cy}^{gt} - b_{cy}^{pred}}{C_h}\right)^2$$

The shape loss $\Omega$ is written as:
$$\Omega = \sum_{t=w,h} (1 - e^{-w_t})^{\theta}$$
$$w_w = \frac{|w - w^{gt}|}{\max(w, w^{gt})}, w_h = \frac{|h - h^{gt}|}{\max(h, h^{gt})}$$

The IOU loss $L_{IoU}$ is calculated using the simple expression:
$$L_{IoU} = 1 - IoU$$
$$IoU = \frac{|B^{pred} \cap B^{gt}|}{|B^{pred} \cup B^{gt}|}$$

Combining all these loss functions, the SIoU loss $L_{SIoU}$ is computed as:
$$L_{SIoU} = L_{IoU} + \frac{\Delta + \Omega}{2}$$

\section{Experiments}
\subsection{Computing Environment and Model Configuration}
In this study, we utilized the openly accessible cloud Jupyter-based platform, Google Colaboratory, which provides access to NVIDIA A100-SXM4 GPU and up to 40 GB of high-memory virtual machines for training the models and making predictions. In addition, the model architecture was developed using Python 3.10 and the PyTorch framework. As seen in Table \ref{tab:Table 1}, the model training configuration utilizes the Stochastic Gradient Descent (SGD) optimizer with initial and final learning rates of $1 \times 10^{-2}$, weight decay of $5 \times 10^{-4}$, and a momentum value 0.937. The model was trained on batches of 16 images per iteration over 100 epochs, keeping all other parameters at their default settings. To understand the effects of different training input image sizes on the performance of the model, we experimented with training input image size of 320 and 640 respectively. Data augmentation techniques and transfer learning strategies were automatically integrated into the training pipeline to further enhance the model's learning.
\begin{table}[ht!]
    \centering
    \caption{The selected model hyperparameters}
    \label{tab:Table 1}
    \begin{tabular}{|c|c|}
        \hline
        \textbf{Parameter} & \textbf{Value} \\
        \hline
        Image size & 320 \& 640 \\
        \hline
        Epochs & 100 \\
        \hline
        Batch size & 16 \\
        \hline
        Momentum & 0.937 \\
        \hline
        Learning rate & 0.01 \\
        \hline
        Optimizer & SGD \\
        \hline
        Activation function & SiLU \\
        \hline
        Copy-paste & 0.3 \\
        \hline
        Mix-up & 0.15 \\
        \hline
        Mosaic & 1.0 \\
        \hline
        Flip Up–Down & 0.5 \\
        \hline
        Flip Left–Right & 0.5 \\
        \hline
        \end{tabular}
\end{table}
\subsection{Dataset}
The dataset used for this work was obtained from two sources:
\begin{itemize}
    \item Training and validation set: \cite{Kaziakhmedov2023} of the NeuroEye Team open-sourced a self-gathered insulator dataset. The data comprises 802 disc-type glass insulator images (640 × 640 pixels) captured from a bottom-up view. The data also include images without corresponding label files, to add them as background images in the training and validation pipelines to reduce the degree of false positives (See Table \ref{tab:Table 2}). An 8:2 random split was adopted to allocate the data into training and validation sets, respectively, with the training images containing over 700 missing insulator instances.
    \item Blind test set: The blind dataset was obtained from the Innopolis High Voltage Challenge hosted on Kaggle (\citealp{Novikov2023}). The data consists of 30 high-resolution (4000 × 2250 pixels) blind test images captured from a top-down perspective by drones flying along the axis of the power line at heights ranging from 15 to 70 meters, with a camera tilt angle ranging from 45° and 70° degrees. To accommodate available computing resources, the images were resized to 3008 × 3008 pixels. 
\end{itemize}
\begin{table}[ht!]
    \centering
    \caption{Data summary}
    \label{tab:Table 2}
    \resizebox{\textwidth}{!}{%
    \begin{tabular}{|c|c|c|c|}
        \hline
         & \textbf{Training set} & \textbf{Validation set} & \textbf{Test set} \\
        \hline
        Image size & 640 x 640 & 640 x 640 & 3008 x 3008 \\
        \hline
        Images with labels & 527 & 132 & 30 \\
        \hline
        Images without labels & 114 & 29 & 0 \\
        \hline
        Total images & 641 & 161 & 30 \\
        \hline
    \end{tabular}%
    }
\end{table}

\subsection{Data Augmentation}
Given the limited number of training images, we augmented the training data to increase the diversity of the dataset and increase the ability of the model to scale across various cases. Some augmentation techniques used involve flipping the image horizontally (left-right) and vertically (up-down). It also incorporates more advanced augmentation methods such as Mosaic, which combines four images into one, new composite image; Mix-up, which generates a weighted combination of random image pairs; and Copy-Paste, which applies random scale jittering to one image and then pastes it onto another. All these augmentation techniques are probability-controlled, indicating the chances of the augmentation being applied in the training pipeline. However, the mosaic augmentation was turned off at the 90th epoch to improve model convergence. It is worth noting that the augmentation methods were not applied to the validation data.
\begin{figure}[htp]
    \centering
    \includegraphics[width=\textwidth]{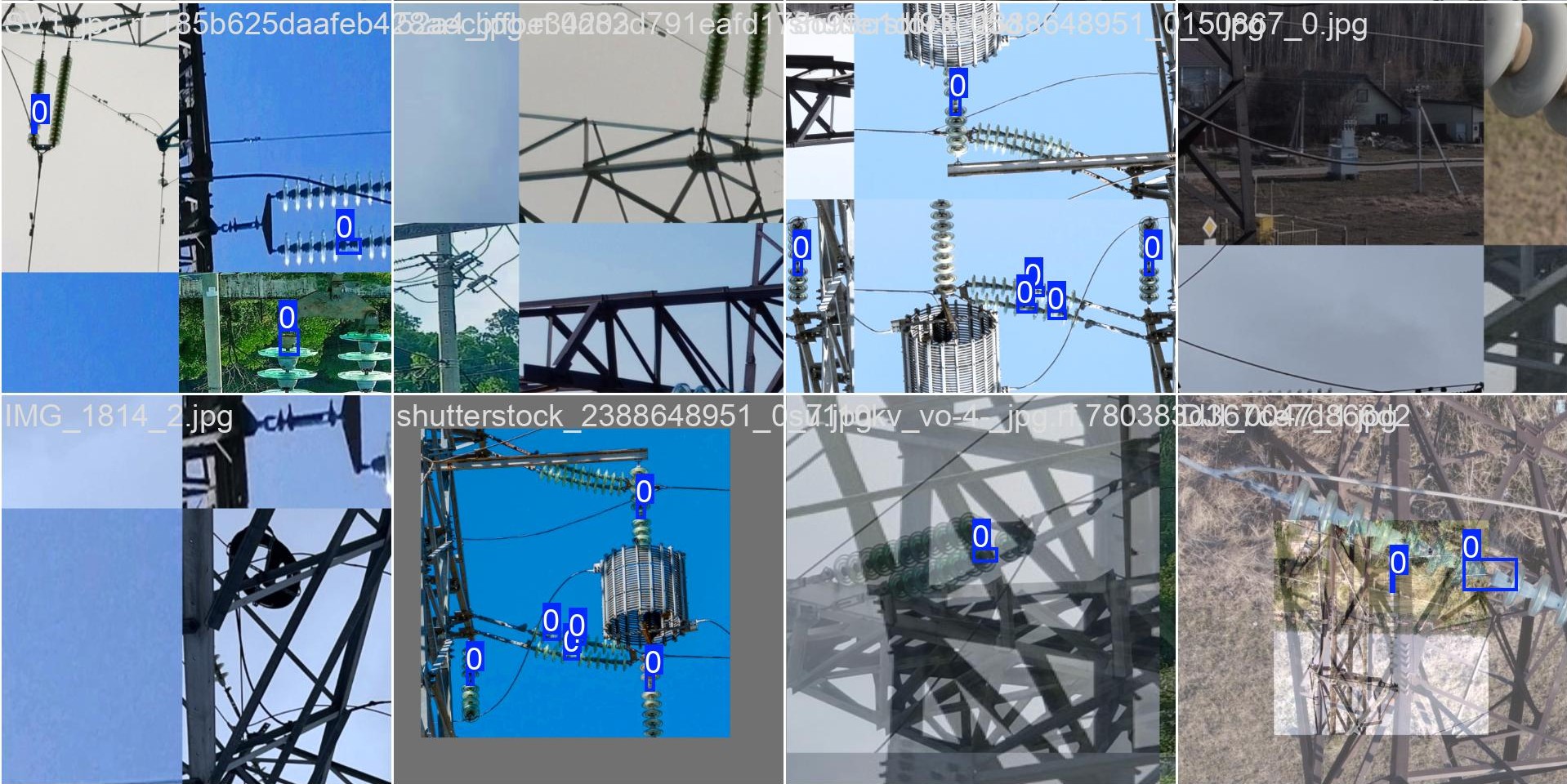}
    \caption{Images showing outcome of data augmentation}
    \label{fig:aug}
\end{figure}

\subsection{Evaluation metrics}
In this work, we consider the application of attention-based YOLOv8 for real-time insulator defect detection. The primary evaluation metric we consider is frames per second (FPS), alongside other metrics like the number of parameters, recall, mean average precision (mAP$_{0.5}$), and giga floating-point operations per second (GFLOPs).\\
\begin{itemize}
    \item \textbf{Parameters (Params)}: The number of parameters depends on a model's complexity, which controls its performance and the computational demand needed for training. The higher the parameters the more complex the model. This leads to better performance but at the expense of higher allocation of computational resources. In real-time detection, it is crucial to balance the number of parameters with the computational cost of inference.
    \item \textbf{Giga Floating Point Operations per Second (GFLOPs)}: GFLOPs is a metric that measures the number of billion floating-point operations a model performs per second and is used to assess a model's complexity and processing speed on hardware systems. In real-time detection tasks, models with lower GFLOPs are the better option for faster execution, although higher GFLOPs can lead to improved accuracy at the cost of increased computational complexity.
    \item \textbf{Frames Per Seconds (FPS)}: The FPS metric is desirable in real-time object detection tasks because it measures the rate at which a model processes images per second. It is controlled by the model complexity and hardware. It is calculated as the inverse of the inference time, typically measured in milliseconds. A model with a higher FPS is preferred for real-time applications as it can handle more frames efficiently, ensuring faster detection or processing.
    \item \textbf{Precision (P), Recall (R) \& Mean Average Precision (mAP$_{0.5}$)}: Precision measures the proportion of true positive predictions made by the model, indicating how accurate the positive predictions are. Recall, on the other hand, evaluates the model's sensitivity to locating true positives. These metrics are combined to calculate the mean Average Precision (mAP$_{0.5}$), which provides an overall measure of the model's performance in detecting insulator defects at an overlapping threshold of 0.5. A higher mAP score indicates better performance in detecting the defects. 
\end{itemize}

\subsection{Experimental Results} 

\noindent As mentioned earlier, we trained our model using training sets with low-resolution input image sizes of 320 and 640, then tested its performance on high-resolution test sets with an input size of 3008. This experimentation aimed to demonstrate the superior performance of our proposed YOLOv8+ELA model compared to other attention-based YOLOv8 models, such as YOLOv8+ECA, YOLOv8+MLCA, YOLOv8+CA, and YOLOv8+CBAM, for insulator defect detection in high-resolution images, despite being trained on lower-resolution input images. \\
\begin{table}[ht!]
\centering
\caption{Performance comparison of different YOLOv8 models on test set: Best performance is highlighted in \textcolor{blue}{Blue} and Second best in \textcolor{red}{Red}. Note that in the P, R, and mAP$_{0.5}$ columns, the upper values represent the performance metrics when the training image size is set to 320, while the lower values correspond to the performance with a training image size of 640.}  
\label{tab:results}
\resizebox{\textwidth}{!}{%
   \begin{tabular}{|c|c|c|c|c|c|c|}
    \hline
    \textbf{Model} & \textbf{Params (M)} & \textbf{P (\%)} & \textbf{R (\%)} & \textbf{mAP$_{0.5}$ (\%)} & \textbf{FPS} & \textbf{GFLOPs} \\
    \hline
    \multirow{2}{*}{YOLOv8} & \multirow{2}{*}{11.13} & 76.1 & 73.2 & 74.5 & \multirow{2}{*}{74.07} & \multirow{2}{*}{28.4} \\
                             &       & 88.5 & 80.5 & 86.0 &       &       \\
    \hline
    \multirow{2}{*}{YOLOv8+ECA} & \multirow{2}{*}{11.13} & 87 & 75.6 & 84.7 & \multirow{2}{*}{72.99} & \multirow{2}{*}{28.4} \\
                             &       & 96.7 & 90.2 & 95.6 &       &       \\
    \hline
    \multirow{2}{*}{YOLOv8+MLCA} & \multirow{2}{*}{11.13} & 89.2 & 75.6 & 84.9 & \multirow{2}{*}{68.97} & \multirow{2}{*}{28.6} \\
                             &       & 96.7 & \textcolor{blue}{\textbf{95.1}} & \textcolor{red}{\textbf{96.9}} &       &       \\
    \hline
    \multirow{2}{*}{YOLOv8+CA} & \multirow{2}{*}{11.16} & 88.2 & 78.0 & 84.3 & \multirow{2}{*}{67.57} & \multirow{2}{*}{28.5} \\
                             &       & 89.7 & 95.1 & 96.2 &       &       \\
    \hline
    \multirow{2}{*}{YOLOv8+CBAM} & \multirow{2}{*}{11.67} & 90.2 & 78.0 & 87.2 & \multirow{2}{*}{69.44} & \multirow{2}{*}{29.2} \\
                             &       & \textcolor{red}{\textbf{97.4}} & 89.7 & 96.4 &       &       \\
    \hline
    \multirow{2}{*}{YOLOv8+ELA (Ours)} & \multirow{2}{*}{11.14} & 96.9 & 77.1 & 89.0 & \multirow{2}{*}{\textcolor{blue}{\textbf{74.63}}} & \multirow{2}{*}{28.5} \\
                             &       & \textcolor{blue}{\textbf{100}} & \textcolor{red}{\textbf{92.6}} & \textcolor{blue}{\textbf{96.9}} &       &       \\
    \hline
\end{tabular}%
}
\end{table}

\noindent Similar to YOLOv8+ELA, attention modules in the other models were also integrated into the neck part of YOLOv8. As seen in Table \ref{tab:results}, the performance of the attention-based models generally improved when trained with an input size of 640 compared to 320. For instance, YOLOv8+ECA achieved mAP$_{0.5}$ scores of 84.7\% and 95.6\% when trained with input sizes of 320 and 640, respectively. YOLOv8+MLCA further enhanced performance, reaching mAP scores of 84.9\% and 96.9\% with the same input sizes. In terms of other metrics like precision and recall, the test precision score when the training image size is 640 is the same for both YOLOv8+ECA and YOLOv8+MLCA, as is the test recall for the 320 training image size. However, YOLOv8+MLCA showed an increase in FLOPs by 0.2G, which resulted in a decrease in detection speed by 4.02 FPS.\\

\noindent The work of \cite{xu2024ela} emphasized that ELA improves upon the CA and CBAM modules. This improvement is evident in insulator defect detection, as YOLOv8+ELA achieved the best test mAP$_{0.5}$ score of 89\% and 96.9\% when trained with input sizes of 320 and 640, respectively — an average improvement of 3.3\% and 0.6\% over the YOLOv8+CA and YOLOv8+CBAM models combined. Although YOLOv8+MLCA and the proposed YOLOv8+ELA model achieve the same mAP$_{0.5}$ score of 96.9\% when trained with an input size of 640, YOLOv8+ELA demonstrates a higher precision value of 100\%, indicating its superior sensitivity to true positive detection. While the YOLOv8+CA and YOLOv8+CBAM models have, on average, 0.28M more learnable parameters than YOLOv8+ELA, they were unable to outperform YOLOv8+ELA in both speed and accuracy across all metrics. It is worth noting that while training on larger input sizes results in an increase in training time, the test inference time remains consistent due to the same high-resolution test image size used across all models. In addition, the varying training image sizes did not impact the models' parameters, FPS, and GLOPs on the test sets.

\subsection{Ablation Study}
\noindent This work aims to improve the baseline YOLOv8 model by incorporating ELA blocks into the neck part and replacing criterion function with SIoU loss. To evaluate the effectiveness of these two enhancement strategies, ablation experiments were conducted using the same model configuration settings, experimental setups, and data conditions. These experiments demonstrate the impact of each improvement on the model's performance.\\

\begin{figure}[htp]
    \centering
    \includegraphics[width=10cm]{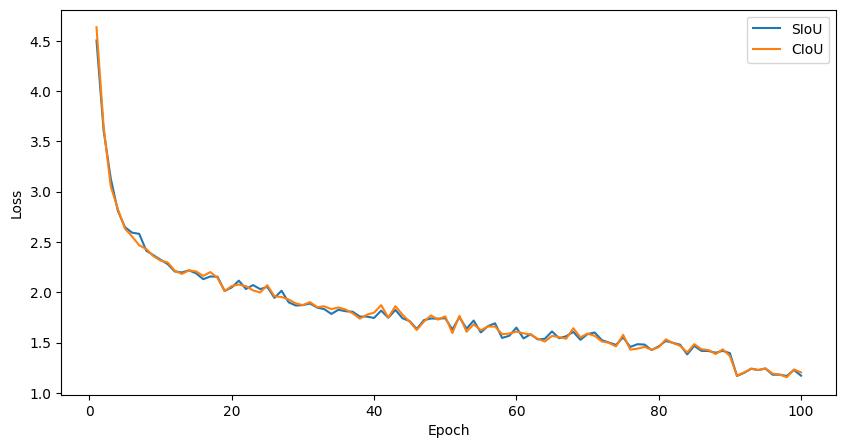}
    \caption{The Comparison between CIoU and SIoU Loss}
    \label{fig:Loss}
\end{figure}  

\noindent Figure \ref{fig:Loss} presents a comparison between the training loss curves of YOLOv8+ELA using SIoU and CIoU loss functions. Visual analysis shows that the training loss decreases with increasing iterations, indicating the model’s adjustment of its weights and parameters to the insulator dataset. The model begins converging at the 5th epoch and continues to decrease consistently throughout the training process. Notably, the model using SIoU starts with a significantly lower initial loss compared to CIoU, and by the final epoch, the SIoU loss remains consistently lower. This reduced loss leads to an overall improvement in model optimization.\\

\noindent Furthermore, the proposed YOLOv8 model with ELA blocks outperforms the original YOLOv8 model across all training input sizes. As shown in Table 4, the mAP$_{0.5}$ of the YOLOv8+ELA model increased by 14.5\% and 10.9\% for training input sizes of 320 and 640, respectively. Despite a slight increase in the number of parameters and FLOPs from 11.13M to 11.14M and 28.4G to 28.5G, respectively, the YOLOv8+ELA model was able to achieve a processing speed of up to 74.63 FPS, representing an improvement of 0.56 FPS compared to the original YOLOv8 model. This highlights that the proposed model will perform better in real-time deployment, offering improvements in both speed and accuracy. \\

\begin{figure}[htbp]
    \centering
    \begin{subfigure}{0.47\textwidth}
        \centering
        \includegraphics[width=\linewidth]{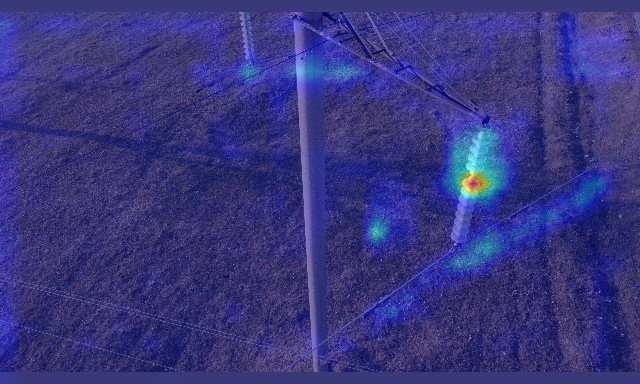}
        \caption{YOLOv8}
    \end{subfigure}
    \hfill
    \begin{subfigure}{0.5\textwidth}
        \centering
        \includegraphics[width=\linewidth]{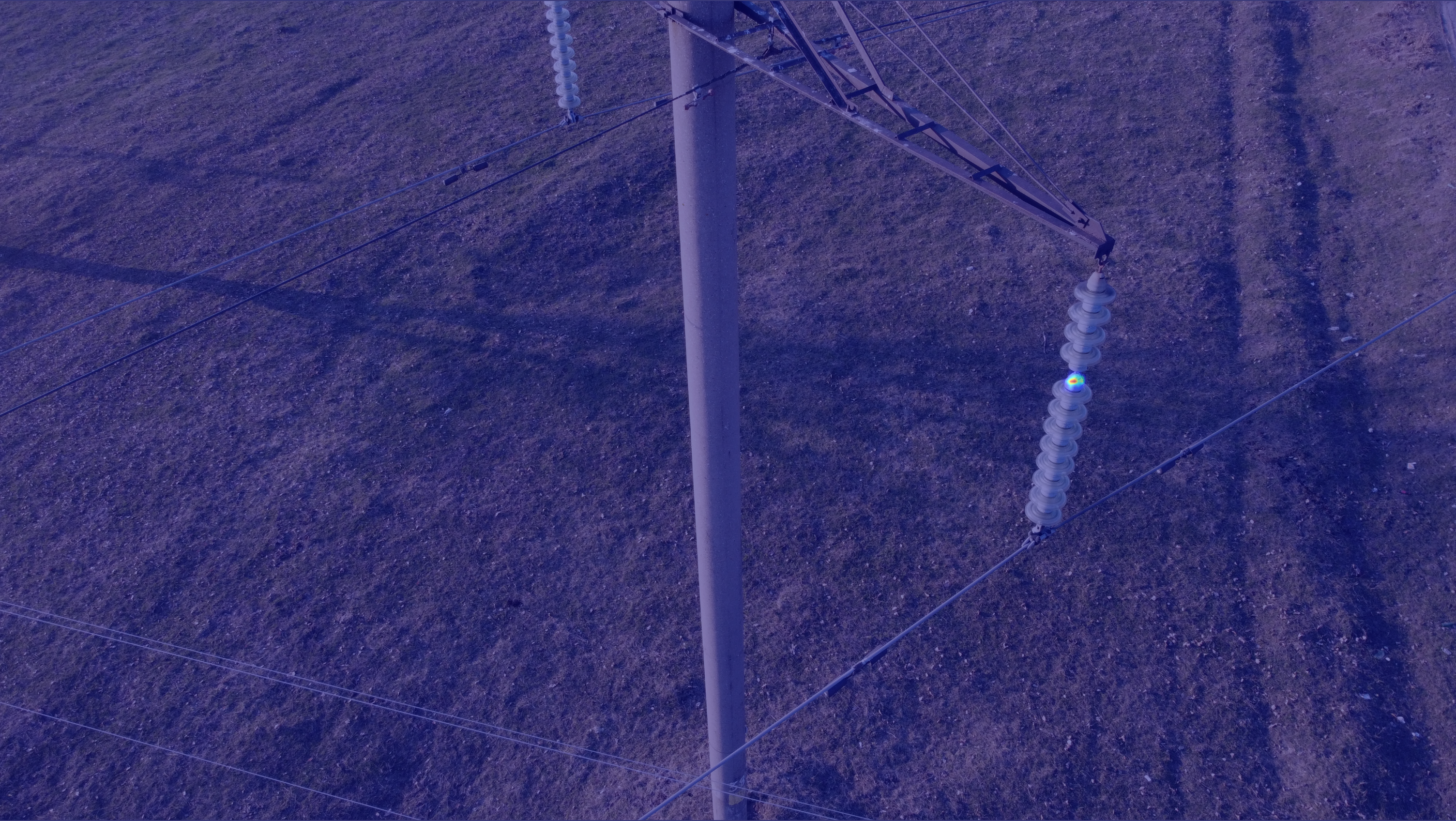}
        \caption{YOLOv8+ELA}
    \end{subfigure}
    \caption{Heatmap of YOLOv8 and YOLOv8+ELA.}
    \label{fig:heat}
\end{figure}

\noindent Figure \ref{fig:heat} depicts the heatmap of the baseline YOLOv8 and YOLOv8+ELA models generated using GradCAM (Gradient-weighted Class Activation Mapping; \citealp{Selvaraju2017}). It is seen that the baseline model, at deeper layers (closer to the output layer), paid specific attention to features of insulators with defects and some background features. However, with YOLOv8+ELA, the background interference is completely removed due to its ability to only learn features related to insulators with defects. This process highlights how the proposed model refines its attention from background features and good insulators to insulators with defects in the power transmission line.\\

\begin{figure}[htbp]
    \centering
    \begin{subfigure}{0.48\textwidth}
        \centering
        \includegraphics[width=\linewidth]{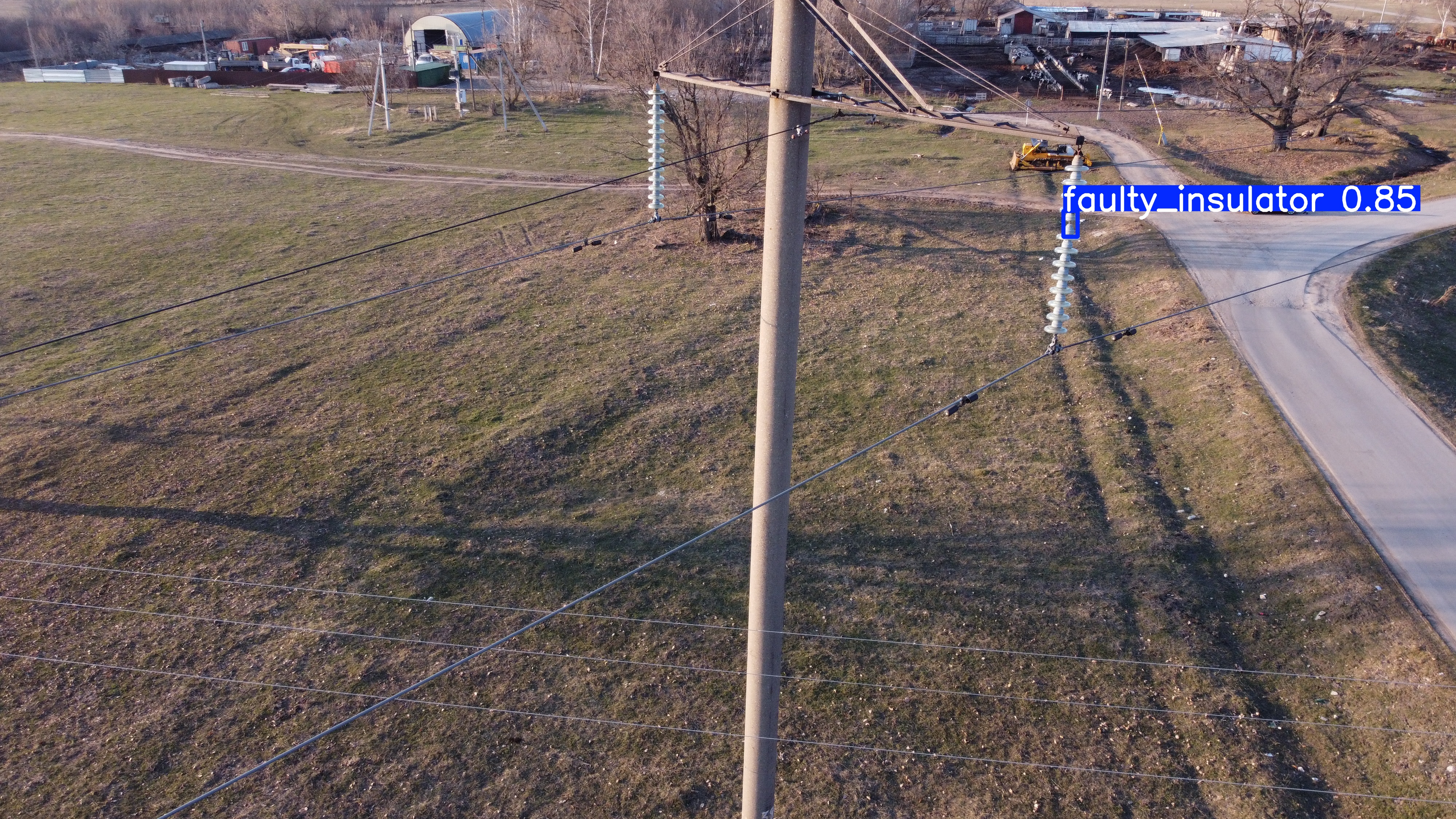}
    \end{subfigure}
    \hfill
    \begin{subfigure}{0.48\textwidth}
        \centering
        \includegraphics[width=\linewidth]{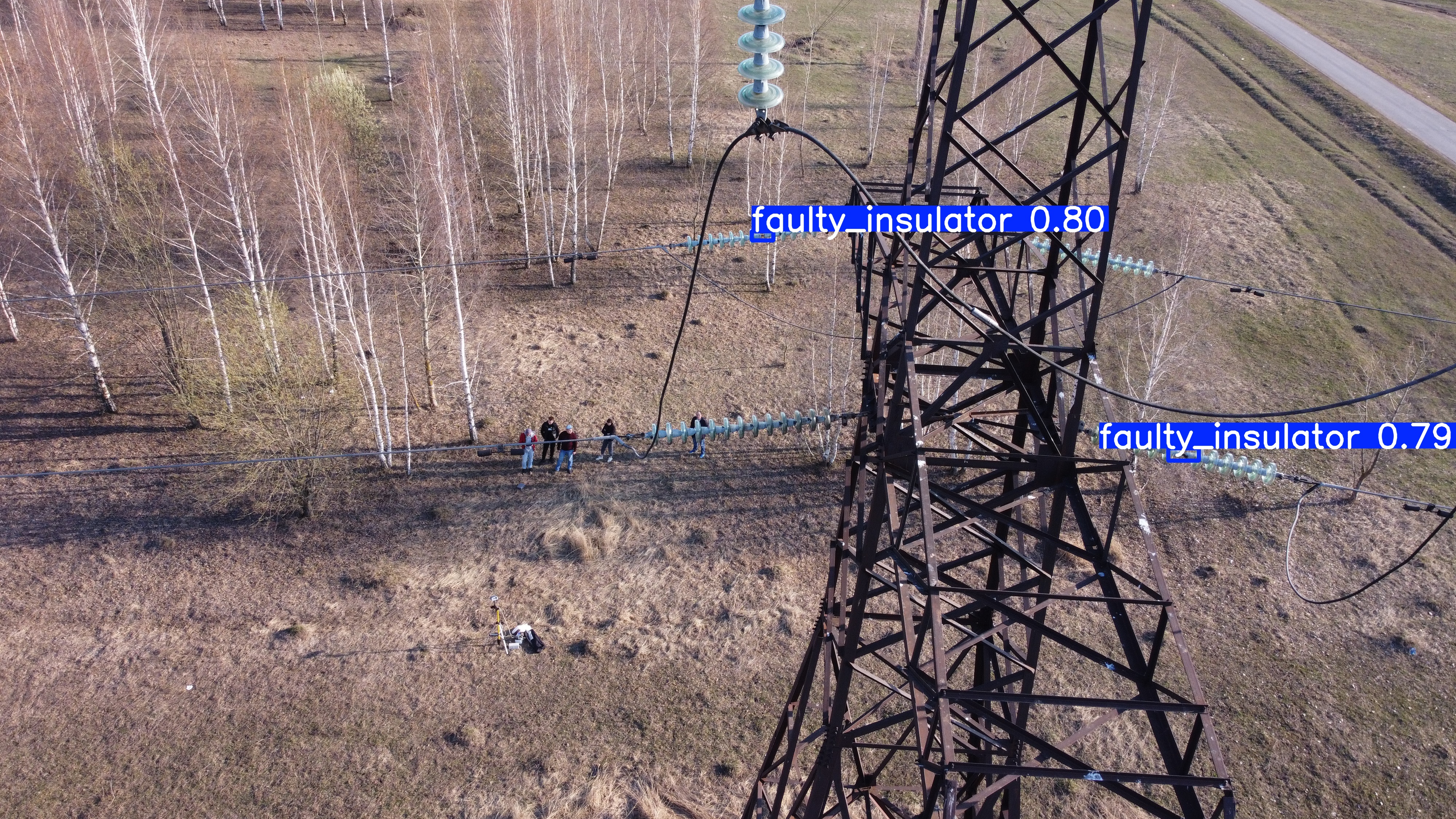}
    \end{subfigure}
    \caption{Some detection results}
    \label{fig:detect}
\end{figure}

\noindent Figure \ref{fig:detect} presents some detection results of YOLOv8+ELA under complex background conditions. The images include various objects such as trees, tower inspection personnel, road networks, cars, and houses. Additionally, the drone's distance from the tower renders the insulators small in the images, making the detection more challenging. Despite these difficulties, YOLOv8+ELA successfully detected all defective insulators with confidence scores of up to 80\%.\\
\section{Conclusion}
\noindent The study successfully demonstrated the integration of deep learning for real-time inspection of power transmission lines using high-resolution UAV images. It also validated that attention-based one-stage YOLOv8 detectors are effective for real-time detection of defective insulators in high-resolution images compared to the baseline model, even when trained on limited and low-resolution datasets. Despite the training set consisting of only images captured from bottom-up angles, the results show that the proposed model effectively processes images taken from top-down perspectives. Notably, our proposed YOLOv8+ELA model achieved the highest performance, with a mAP$_{0.5}$ score of up to 96.9\% and FPS of 74.63. With this, the YOLOv8+ELA model proves to be a robust solution for real-time insulator defect detection in UAV imagery, combining high precision, recall, and mAP performance. The model’s ability to generalize across different perspectives and resolutions makes it an ideal candidate for scalable deployment in power line inspections. Furthermore, its consistent performance under challenging environmental conditions, such as complex backgrounds and small object detection, underscores its potential for improving the efficiency and accuracy of transmission line maintenance, ultimately contributing to more reliable power delivery systems.

\bibliography{References}
\hypertarget{myrefsection}{}
\begingroup
\renewcommand{\section}[2]{}
\bibliographystyle{apalike}

\end{document}